\pdfoutput=1

\documentclass[11pt]{article}

\usepackage[]{EMNLP2023}

\usepackage{times}
\usepackage{latexsym}

\usepackage[T1]{fontenc}

\usepackage[utf8]{inputenc}

\usepackage{microtype}

\usepackage{inconsolata}

\usepackage{url}
\usepackage{subfigure}
\usepackage{multirow}
\usepackage{enumitem}
\usepackage{stmaryrd}
\usepackage{array}
\usepackage{threeparttable}
\usepackage{blindtext}
\usepackage{amssymb}
\usepackage{CJKutf8}
\usepackage[ruled,vlined,linesnumbered,longend]{algorithm2e}
\usepackage{svg}
\usepackage{placeins}

\usepackage{soul}
\usepackage[utf8]{inputenc}
\usepackage{graphicx}
\usepackage{amsmath}
\usepackage{booktabs}
\usepackage{lipsum}

\usepackage{threeparttable}
\usepackage{makecell}
\usepackage{booktabs}
\usepackage{multirow}
\usepackage{marvosym}
\usepackage{tikz}
\newcommand{\hide}[1]{}

\newcommand{\model}{ChatGPT}

\usepackage{amsmath,amsfonts,bm}

\def\eqref#1{equation~\ref{#1}}

\def\1{\bm{1}}

\DeclareMathAlphabet{\mathsfit}{\encodingdefault}{\sfdefault}{m}{sl}
\SetMathAlphabet{\mathsfit}{bold}{\encodingdefault}{\sfdefault}{bx}{n}

\title{ChatGPT is a Potential Zero-Shot Dependency Parser}

\author{
    Boda Lin$^{1\ddagger}$, Xinyi Zhou$^{2\ddagger}$, Binghao Tang$^1$, Xiaocheng Gong$^1$, Si Li$^{1*}$ \\
    $^1$School of Artificial Intelligence, Beijing University of Posts and Telecommunications\\
    $^2$Department of Chinese Language and Literature, East China Normal University\\
  \texttt{\{linboda, lisi\}@bupt.edu.cn} \\
   }
\begin{document}
\maketitle

\begin{abstract}
Pre-trained language models have been widely used in dependency parsing task and have achieved significant improvements in parser performance. 
However, it remains an understudied question whether pre-trained language models can spontaneously exhibit the ability of dependency parsing without introducing additional parser structure in the zero-shot scenario. 
In this paper, we propose to explore the dependency parsing ability of large language models such as ChatGPT and conduct linguistic analysis.
The experimental results demonstrate that ChatGPT is a potential zero-shot dependency parser, and the linguistic analysis also shows some unique preferences in parsing outputs.
\let\thefootnote\relax\footnotetext{$^\ddagger$Boda Lin and Xinyi Zhou make equal contribution}
\let\thefootnote\relax\footnotetext{$^*$Corresponding author}
\end{abstract}

\section{Introduction}
Dependency parsing is a fundamental task in Natural Language Processing and have many applications in downstream tasks, such as machine translation~\cite{bugliarello2020mt}, 
question answering~\cite{teney2017qa}, 
and information retrieval~\cite{chandurkar2017ir}.
Previous research mainly focus on how to design the parser structure and parsing algorithms to achieve better performance in different scenarios~\cite{dozat2017biaffine,ma2018stackptr,li2019codt}.

The linguistic base of dependency parsing is the dependency grammar~\cite{jarvinen1998towards}, which come from the linguists' research about linguistic rules and phenomena.
Notably, Pre-trained Language Models~(PLMs) can also be viewed as "linguists" that automatically learn rules from a vast amount of natural language texts.
Therefore, investigating whether these PLMs spontaneously learn certain syntactic rules during the pre-training stage is a valuable research topic.

After the proposal of the BERT model~\cite{devlin2019bert}, numerous probing studies explore the specific functions learned by each layer of BERT, which also touches upon the research of self-acquisition of syntax for PLMs~\cite{rogers-etal-2020-primer}.
However, the BERT model presents the following limitations:
1) 
These studies often induce dependency parsing results from the attention mechanism of BERT.
These parsing results do not stem from natural generative steps. 
In particular, parameterized probing methods could introduce external information interference~\cite{wu-etal-2020-perturbed}; 
2) Due to model limitations, these works are often confined to relatively simple datasets and struggles to directly yield syntactic results with dependency relation labels. 
Despite the subsequent BART~\cite{lewis2020bart}, T5~\cite{colin2020t5}, and other encoder-decoder structured PLMs showing good performance in language generation, the complexity of the expression form inherent in the task of dependency parsing still makes it challenging to induce syntactic results from such PLMs in a more straightforward manner.

Recently, Large Language Models~(LLMs) such as InstructGPT~\cite{ouyang2022training} and ChatGPT~\footnote{https://openai.com/blog/chatgpt}, which possess superior generative capabilities, have been introduced in NLP. 
These models have achieved impressive performance on various NLP tasks, including question answering, reading comprehension, and summarization~\cite{ouyang2022training}, even in a zero-shot fashion, providing a crucial key to investigating the innate syntactic abilities of language models. 
We are interested in the following questions: 1) Do LLMs like ChatGPT possess zero-shot dependency parsing capabilities? 
2) If so, do the outputs of ChatGPT still maintain a similar structure for similar sentences even between different languages?
3) Further more, do these parsing results contain some preferences that can be summarized?
\begin{figure*}[t]
    \centering
    \includegraphics[width=1.0\linewidth]{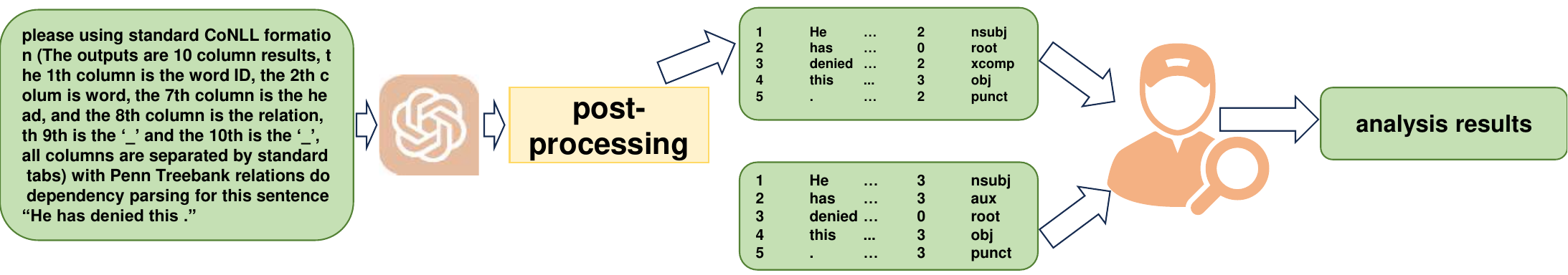}
    \caption{The total framework of our work.}
    \label{fig:framkework}
\end{figure*}

In this paper, we explore using ChatGPT and other LLMs to achieve zero-shot dependency parsing, and conduct linguistic analysis on these parsing results to answer these questions.

The results demonstrate the ChatGPT is a potential zero-shot dependency parser and the outputs of ChatGPT maintain similar structure in different languages.
And we summarize some parsing preferences of ChatGPT through linguistic analysis. 
The most surprising finding is that in some cases, ChatGPT outputs are more in line with linguistic rules than gold annotations.


\section{Related Work}
\subsection{Dependency Parsing}
PLMs are widely used in previous dependency parsing research.
Biaffine~\cite{dozat2017biaffine} with BERT~\cite{devlin2019bert} is a very simple and effective parser.
\citet{yang-tu-2022-headed} proposes a graph-based parser based on headed spans.
\citet{lin-etal-2022-dpsg} design the serialization of parsing trees and enabling the T5 model directly generate parsing sequence.
Besides, the newest state-of-the-art model Hexatagging~\cite{amini2023hexatagging} cast the parsing tree into the hexatag sequence and also only use BERT to achieve parsing.
But these methods still rely on the supervised learning paradigm.


\subsection{Probing}
The probe research of the BERT model can be roughly divided into two categories: parametric methods and non-parametric methods.
\citet{jawahar-etal-2019-bert} use a series of probing tasks to indicate that while BERT does capture some of this information, it is not always explicitly encoded within BERT's representations.
\citet{wu-etal-2020-perturbed} use non-parametric method to achieve probing task for BERT.

\section{ChatGPT Parsing}
As shown in Figure~\ref{fig:framkework}, our approach leverages the prompt to enable ChatGPT for zero-shot dependency parsing.
Specifically, we define the input format as a sentence adhering to the original word segmentation of the parsing corpus, while the output is generated in CoNLL format.


The parsing results from ChatGPT may exhibit various format-related issues, which include word missing, format disruption, word segmentation, word scrambling, and multiple outputs.
We use post-processing to filter the illegal outputs, the specific statistical details can refer to the Appendix~\ref{app:post}.








\section{The Consistency}
In order to explore whether the dependency parsing outputs of ChatGPT maintain the consistency in different languages, we conduct corresponding experiments and use Dependency Tree Edit Distance~(DTED)~\cite{mccaffery-nederhof-2016-dted} to measure the similarity between dependency syntax trees in different languages.

\begin{equation}
    DTED(T_a, T_b) = 1 - \frac{EditDist(T_a, T_b)}{\max(|T_a|, |T_b|)}
\end{equation}
Where $T_a$ and $T_b$ means two parsing trees from different language, and the value of DTED score is from [0, 1].
The EditDist is calculate based on the Tree Edit Distance algorithm~\cite{zhang1989simple}.

\section{Parsing Ability Experiment}
\subsection{Setting}
In this paper, we investigate the parsing ability of ChatGPT from two different settings:
1) The normal zero-shot parsing setting. In this setting, we use gpt-3.5-turbo, HuggingChat\footnote{\url{https://huggingface.co/chat}}, Vicuna-13B\footnote{\url{https://lmsys.org/blog/2023-03-30-vicuna}} and ChatGLM-6B~\cite{zeng2022glm} to directly achieve dependency parsing on English and Chinese.
2) The cross-lingual parsing setting.
In this setting, we collect the sentence pairs from English and Chinese which have the similar parsing structure.
Then we conduct dependency parsing on these sentence pairs and analyze the same or difference between English and Chinese.
In order to avoid the influence of generating randomness, we set the temperature of ChatGPT to 0 in all experiments.

\subsection{Dataset}
For English parsing, we choose the proverbial benchmark Penn Treebank~(PTB)~\cite{marcus1993ptb3}, Chinese Treebank 
 5 (CTB5)~\cite{xuectb5} and 12 languages from Universal Dependencies Version 2.2~(UD2.2)~\cite{nivre2016ud} following the previous work~\cite{ma2018stackptr}.
For PTB and CTB5, we follow \citet{ma2018stackptr} to use the Stanford basic Dependencies representation~\cite{de2006sd} of PTB and CTB converted by Stanford parser\footnote{http://nlp.stanford.edu/software/lex-parser.html}. 

For the cross-lingual parsing setting, we use the DTED score to choose the most similarity top-50 sentences from en-ewt-test and zh-cfl-test of UD 2.2.

\subsection{Parsing Ablility}
\begin{table}[t]
    \centering
    \begin{tabular}{cccc}
    \toprule
    \textbf{Dataset} & \textbf{Method} & \textbf{UAS} & \textbf{LAS} \\
    \midrule
    \multirow{5}{*}{PTB} & Biaffine & $95.74$ & $94.08$ \\
     & StackPTR & $95.87$ & $94.19$ \\
     & DPSG & $96.64$ & $95.82$ \\
     & Hexatagging & $\textbf{97.40}$ & $\textbf{96.40}$ \\
     & ChatGPT & $40.22$ & $28.61$ \\
     \midrule
    \multirow{4}{*}{CTB5} & Biaffine & $89.30$ & $88.23$ \\
     & StackPTR & $90.59$ & $89.29$ \\
     & Hexatagging & $\textbf{93.20}$ & $\textbf{91.90}$ \\
     & ChatGPT & $28.08$ & $11.93$ \\
     \midrule
     en-top & ChatGPT & $32.88$ & $26.68$ \\
     zh-top & ChatGPT & $53.19$ & $39.36$ \\
     \bottomrule
    \end{tabular}
    \caption{The results of ChatGPT and other traditional supervised parsing methods on PTB and CTB5. The en-top and zh-top mean the top 100 sentences extract from the en-ewt-test and zh-cfl-test.}
    \label{tab:main_results}
\end{table}
\begin{table}[t]
    \centering
    \begin{tabular}{ccc}
    \toprule
         & Gold & ChatGPT \\
         \midrule
      Avg\_DTED & 0.64 & 0.45 \\
    \bottomrule
    \end{tabular}
    \caption{The average DTED score between top 50 similarity sentences of English and Chinese.}
    \label{tab:dted}
\end{table}
\begin{table*}[t]
    \centering
    \scalebox{0.8}{
    \begin{tabular}{c|c|c|c|c|c|c|c|c|c|c|c|c|c}
    \toprule
    \textbf{Model} & \textbf{bg} & \textbf{ca} & \textbf{cs} & \textbf{de} & \textbf{en} & \textbf{ es} & \textbf{fr} & \textbf{it} & \textbf{nl} & \textbf{no} & \textbf{ro} & \textbf{ru} & \textbf{AVG}\\
    \midrule
    Biaffine & $90.30$ & $94.49$&  $92.65$&  $85.98$ &$ 91.13$&  $93.78$ & $91.77$&  $94.72$ & $91.04 $& $94.21$ & $87.24$ & $94.53$ & $91.82$ \\
    StackPTR & $89.96$ & $92.39$ & $90.94$& $86.16$&  $89.83$ & $91.52$& $89.88$ &$92.55$&  $91.73$ &$93.62$&  $85.34$&  $93.07$ & $90.58$\\
     DPSG & $\textbf{93.92}$ &$93.75$ &$\textbf{92.97}$& $84.84$ &$\textbf{91.49}$& $92.37$ &$90.73$ &$94.59$& $\textbf{92.03}$& $\textbf{95.30}$ &$\textbf{88.76}$ &$\textbf{95.25}$ &$\textbf{92.17}$ \\
     Hexatagger & $92.87$ &$\textbf{93.79}$& $92.82$ &$\textbf{85.18}$ &$90.85$& $\textbf{93.17}$ &$\textbf{91.50}$ &$\textbf{94.72}$& $91.89$& $93.95$& $87.54$ &$94.03$& $91.86$ \\
    \midrule
    \model & $35.87$ & $34.04$ & $33.37$ & $26.88$ & $32.63$ & $35.01$ & $29.56$ & $24.60$ & $29.79$ & $30.86$ & $33.78$ & $34.97$ & $31.78$ \\
    \bottomrule
    \end{tabular}
    }
    \caption{Results on 12 languages of UD2.2 in terms of LAS.}
    \label{tab:ud:result}
\end{table*}

According the results shown in Table~\ref{tab:main_results} and Table~\ref{tab:ud:result}, we can answer the first question in the introduction.

There is no doubt that ChatGPT has the ablility of zero-shot dependency parsing.
In fact, this capability is already quite rare in other LLMs. 
We also conducted zero-shot experiments on some other popular LLMs (HuggingChat, Vicuna-13B, ChatGLM-6B)\footnote{Since other LLMs lacking the CoNLL or sequence parsing output capability, we cannot calculate the performance} and found that maintaining a generally correct CoNLL format output is a very challenging task for these LLMs. 
Furthermore, we conducted one-shot demonstration learning experiments for these LLMs, with only Vicuna able to learn a fairly close format through example. 
To rule out the complexity of the CoNLL format itself, we converted the CoNLL format parsing tree into a sequence format following~\cite{lin-etal-2022-dpsg}, but other LLMs were still unable to produce satisfactory results.

The results on Table~\ref{tab:dted} answer the second question in the introduction, the output similarity of ChatGPT between English and Chinese reaches 0.46, which is 72\% of the gold similarity, which shows that for sentences with similar structures in different languages, the output results given by ChatGPT still have a high structural similarity.
We show more details in Appendix~\ref{app:similarity}.
\section{Linguistic Analysis}
In this section, we answer the last question in the introduction.
We analyze the 50 sentences in the test set of en-ewt-test in UD2.2 for English and the total test set of CTB5 for Chinese.
More details and examples are listed in the Appendix~\ref{app:en:example} and Appendix~\ref{app:zh:example}.



\subsection{English Analysis}
More precisely, the analysis conducted on English outputs from ChatGPT reveals distinct error patterns that can be classified into the following:

\textbf{Predicative Verb-Centrism}: 
This phenomenon pertains to the inclination of perceiving the verb in the predicate as the central element. 
This tendency is commonly observed during the processing of clauses, particularly when the subject is omitted.

\textbf{Noun Subject Preference}: 
This tendency reflects a predisposition to designate a noun as the subject within a phrase.

\textbf{Preposition-Case Ambiguity}: This phenomenon pertains to the inconsistent categorization of simple prepositions such as "to", "in", "case", and "for", where they are labeled inconsistently as either prepositions or case.

\textbf{Auxiliary Verb Mislabeling}: 
This phenomenon arises when a sentence comprises both auxiliary verb and non-finite verb. In such cases, ChatGPT tends to designate the auxiliary verb as the root due to the influence of positional placement.

\textbf{Adjective Modifier Ambiguity}: 
This phenomenon emerges when a noun is preceded by multiple modifiers, creating ambiguity in discerning the appropriate modifier.










\subsection{Chinese Analysis}
We also conduct linguistic analysis for CTB5, and the observations are summarized into the following, more statistics details are shown in Appendix~\ref{app:statis.ctb5}.

\textbf{Obj-Dobj}: 
The most prevalent disparity observed between the ChatGPT and the golds is the "obj-dobj". 
Although both "obj" and "dobj" signify a direct object relationship, their usage is inconsistent across the two tagging sets, indicating potential variations in labeling conventions.

\textbf{Compound-Nmod}: 
The differentiation between the "compound" (indicating a compound noun, adjective, or adverb) and "nmod" (representing a nominal modifier of a noun or pronoun) frequently leads to discrepancies between ChatGPT and the golds. 

\textbf{Quote Root}: 
When a sentence portrays a statement attributed to an individual, with the speaker positioned at the end of the sentence, ChatGPT demonstrates a tendency to designate the verb within the quotation as the root, whereas the golds mark the verb "say" as the root.

\textbf{Error Root}: 
ChatGPT occasionally misclassifies punctuation marks as the root, thereby assigning them undue prominence. Additionally, it displays inconsistencies in labeling the main verb as the root, often assigning this role to nouns instead. In certain cases, ChatGPT fails to identify the root.

\textbf{Dobj-Nmod}: 
In certain instances, ChatGPT erroneously labels "dobj" as "nmod". This primarily transpires when there are disagreements regarding the appropriate word to which the dependent should be linked.

\textbf{Nummod-Dep}: 
The utilization of numerals within ChatGPT outputs is typically labeled as "nummod", whereas the gold standard annotations often designate it as "dep". 
It is worth noting that in certain cases, the gold annotation may be inaccurate, incorrectly assuming that the numeral modifies a noun that it logically does not modify.

\subsection{Summarization}


In general, the inconsistencies between ChatGPT and golds can be categorized into three groups: 1) Instances where ChatGPT outputs are incorrect; 2) Cases where ChatGPT and golds can both be considered correct, but differ due to distinct perspectives considered in the annotations; 3) Situations where ChatGPT exhibits greater linguistic normativity compared to the golds.

The third category is intriguing as it highlights the potential of ChatGPT. While traditional parsing methods based on supervised learning achieve impressive performance, the parsing capabilities of these models are constrained by the labeled data.
However, the parsing abilities of ChatGPT acquired through pre-training may transcend this limitation and offer researchers a novel perspective. 
This breakthrough could potentially provide researchers with valuable insights or alternative viewpoints.

\section{Conclusion}

We employ the prompt to investigate the zero-shot dependency parsing capability of ChatGPT and other Language Models (LLMs) on proverbial benchmarks. 
The experimental results substantiate that ChatGPT exhibits promising potential as a zero-shot dependency parser. 
Furthermore, cross-lingual experiments demonstrate the ability of ChatGPT to maintain similarity in parsing outputs across different languages.
Additionally, linguistic analysis is performed to discern the parsing output preferences of ChatGPT. 
The analysis reveals that ChatGPT has the ability to surpass limitations stemming from errors in labeled data. 
\section*{Limitations}


Considering the powerful learning ability of Large Language Models~(LLMs), we use prompt-based method to analyze the zero-shot ability of Dependency Parsing of LLMs.
The different formats of prompts might significantly affect the final outputs and could be a disturbance for our experiments.
Moreover, the datasets and corpora usage of LLMs is unclear and that might influence our linguistic analyses.
In addition, linguistic analysis may be mixed with some subjective judgments, and due to the complex format of the parsing data, many linguistic analysis phenomena are difficult to directly perform data statistics.

\section*{Ethics Statement}
We affirm that our work here does not exacerbate the biases already inherent in the large language models and our linguistic analyses are also only based on those model outputs. As a result, we anticipate no ethical concerns associated with this research.

\bibliography{anthology}
\bibliographystyle{acl_natbib}

\appendix
\section{Post-processing Details}
\label{app:post}
Since the dependency parsing task is a fine-grained task, it has high requirements on vocabulary and output format, and the uncontrollability of LLM itself, there will be many formal errors in the outputs of ChatGPT, as follows:

Word filtering: Since some of the parsing corpora come from political news, certain vocabulary may trigger the filtering policy of ChatGPT, leading to the omission of sensitive words in the output CoNLL results.

Format disruption: Occasionally, ChatGPT may not output in the standard CoNLL format, causing issues such as missing columns, extra columns, or disordered columns.

Word segmentation disruption: This phenomenon is particularly common in languages that require word segmentation, such as Chinese. Even though we clearly pre-segmented the input with spaces, ChatGPT may sometimes employ its own segmentation.

Word omission: In lengthy sentences, there might be instances where a sequence of words is missing.

Word scrambling: In extended sentences, the outputted CoNLL results may contain parts where the vocabulary is scrambled.

Multiple outputs: In some cases, ChatGPT will give duplicate parsing outputs for a sentence.

Since these formal errors will cause predict and gold to fail to achieve alignment, we use post-processing to filter out the output containing these errors.
The size of the original test sets and the size of the data obtained after post-processing are shown in the Table~\ref{tab:ud:statistics}.

\begin{table*}[t]
    \centering
    \scalebox{0.78}{
    \begin{tabular}{c|c|c|c|c|c|c|c|c|c|c|c|c|c|c}
    \toprule
     & \textbf{PTB} & \textbf{CTB5} & \textbf{bg} & \textbf{ca} & \textbf{cs} & \textbf{de} & \textbf{en} & \textbf{es} & \textbf{fr} & \textbf{it} &\textbf{ nl} & \textbf{no} & \textbf{ro} & \textbf{ru}\\
    \midrule
    Total Number & $2,416$ & $1,915$ & $1,116$ & $1,846$ & $12,203$ & $977$ & $2,077$ & $2,174$ & $416$ & $482$ & $1,396$ & $3,450$ & $729$ & $6,491$ \\
    Final Number & $1,394$ & $990$ & $518$ & $1,307$ & $8,040$ & $505$ & $1,164$ & $1,644$ & $283$ & $374$ & $614$ & $2,492$ & $498$ & $5,742$ \\
    \bottomrule
    \end{tabular}
    }
    \caption{The number of sentences in the original test set of PTB, CTB5 and 12 languages on UD2.2 and the number of sentences retained after post-processing.}
    \label{tab:ud:statistics}
\end{table*}
\begin{table*}[t]
    \centering
    \scalebox{0.75}{
    \begin{tabular}{cccccc|ccccc}
    \toprule
    \textbf{ID} & \textbf{Word} & \textbf{Pred-Head} & \textbf{Pred-Rel} & \textbf{Gold-Head} & \textbf{Gold-Rel} & \textbf{Word} & \textbf{Pred-Head} & \textbf{Pred-Rel} & \textbf{Gold-Head}& \textbf{Gold-Rel} \\
    \midrule
1 & He & 2&nsubj&3&nsubj & \begin{CJK*}{UTF8}{gbsn}我们\end{CJK*}&2&nsubj&3&	nsubj  \\ 
2 & has & 0&root&3&aux &  \begin{CJK*}{UTF8}{gbsn}要\end{CJK*}&0&root&3	&aux\\
3 & denied &2&xcomp&0&root & \begin{CJK*}{UTF8}{gbsn}去\end{CJK*}&2&xcomp&0	&root \\
4 & this & 3&obj&3&obj & \begin{CJK*}{UTF8}{gbsn}目的地\end{CJK*}&3	&obj&3&obj\\
5 & . & 2&punct&3&punct &  \begin{CJK*}{UTF8}{gbsn}！\end{CJK*}&2&punct&3&punct\\
    \midrule
1 &you& 2&nsubj&3&nsubj & \begin{CJK*}{UTF8}{gbsn}你\end{CJK*}&2&nsubj&3&	nsubj  \\ 
2 & r & 0&root&3&cop &  \begin{CJK*}{UTF8}{gbsn}是\end{CJK*}&0&root&3	&cop\\
3 & retarded &2&xcomp&0&root & \begin{CJK*}{UTF8}{gbsn}学生\end{CJK*}&2&attr&0&root \\
4 & . & 2&punct&3&punct &  \begin{CJK*}{UTF8}{gbsn}？\end{CJK*}&2&punct&3&punct\\
    \midrule
1 &Just& 2&advmod&3&advmod & \begin{CJK*}{UTF8}{gbsn}有点儿\end{CJK*}&2&amod&2&	advmod  \\ 
2 & our & 3&amod&3&nmod:poss &  \begin{CJK*}{UTF8}{gbsn}恼火\end{CJK*}&0&root&0&root\\
3 & standard &0&root&0&root & \begin{CJK*}{UTF8}{gbsn}了\end{CJK*}&2&mark&2&discourse:sp \\
4 & . & 3&punct&3&punct &  \begin{CJK*}{UTF8}{gbsn}。\end{CJK*}&2&punct&2&punct\\
 \bottomrule
    \end{tabular}
    }
    \caption{Three similarity parsing tree pair examples.}
    \label{tab:top}
\end{table*}
\begin{table}[t]
    \centering
    \scalebox{0.65}{
    \begin{tabular}{cccccc}
    \toprule
    \textbf{ID} & \textbf{Word} & \textbf{Pred-Head} & \textbf{Pred-Rel} & \textbf{Gold-Head} & \textbf{Gold-Rel} \\
    \midrule
    \multicolumn{3}{c}{\textbf{Predicative Verb-Centrism}}\\
    1 & What & 4 & nsubj & 0 & root \\
    2 & if & 4 & mark & 4 & mark \\
    3 & Google & 4 & nsubj & 4 & nsubj \\ 
    4 & Morphed & 0 & root & 1 & advcl \\
    5 & Into & 4 & prep & 6 & case \\
    6 & GoogleOS & 5 & pobj & 4 & obl \\
    7 & ? & 4 & punct & 4 & punct \\
    \midrule
    \multicolumn{3}{c}{\textbf{Noun Subject Preference}}\\
    1 & One & 2	&nummod & 5&nsubj \\
    2 & of & 4&	case & 4&case \\
    3 & the & 4	&det & 4 & det \\ 
    4 & pictures & 5&nsubj & 1 & nmod \\
    5 & shows & 0&root & 0&root \\
    6 & a & 7&det & 7&det \\
    7 & flag & 5&obj & 5&obj \\
    8 & that & 9&nsubj:pass & 10&nsubjpass \\
    9 & was & 5 & acl:pass & 10 & aux:pass \\
    10 & found & 9 & auxpass & 7 & acl:relcl \\
    11 & in & 12 & case & 12 & case \\
    12 & Fallujah & 10 & obl & 10 & obl \\
    13 & . & 5 & punct & 5 & punct \\
    \midrule
    \multicolumn{3}{c}{\textbf{Preposition-Case Ambiguity}}\\
    1 & Compare & 0&root & 0&root \\
    2 & the & 3&det & 3&det \\
    3 & flags & 1 & obj & 1 & obj \\ 
    4 & to & 1&prep & 7 & case \\
    5 & the & 6&det & 7&det \\
    6 & Fallujah & 4&pobj & 7&compound \\
    7 & one & 1&dobj & 7&obl \\
    8 & . & 1&punct & 1&punct \\
    \midrule
    \multicolumn{3}{c}{\textbf{Auxiliary Verb Mislabeling}}\\
    1 & It & 2	&nsubj & 3&expl \\
    2 & does & 0&root & 3&aux \\
    3 & seem & 2&ccomp & 0 & root \\ 
    4 & that & 3&mark &7&mark \\
    5 & Iranians & 6&nsubj & 7&nsubj \\
    6 & frequently & 3&advmod & 7&advmod \\
    7 & make & 6&conj & 3&ccomp \\
    8 & statements & 7&dobj & 7&obj \\
    9 & and & 7 &cc & 11&cc \\
    10 & then & 11 & advmod &11&advmod \\
    11 & hide & 7 & conj & 7&conj \\
    12 & behind & 11 & prep & 13&case \\
    13 & lack & 14 & compound & 11&obl \\
    14 & of & 12 & pobj &15&case \\
    15 & proof & 14 & nmod & 13&nmod \\
    16 & . & 2 & punct & 3&punct \\
    \midrule
    \multicolumn{3}{c}{\textbf{Adjective Modifier Ambiguity}}\\
    1 & The & 2&det & 2&det \\
    2 & clerics & 4&nsubj & 3&nsubj \\
    3 & demanded & 4&aux & 0 & root \\ 
    4 & talks & 0 & root & 3 & obj \\
    5 & with & 6&case & 8&case \\
    6 & local & 7&amod & 8&amod \\
    7 & US & 8&compound & 8&compound \\
    8 & commanders & 4&obl & 4&nmod \\
    9 & . & 4&punct & 3&punct \\
    \bottomrule
    \end{tabular}
    }
    \caption{The analysis examples in English.}
    \label{tab:en.ex1}
\end{table}

\section{Examples of Similarity Trees}
\label{app:similarity}
In addition to the calculated DTED scores as shown in Table~\ref{tab:dted} in Section 5, we can also intuitively see from the Table~\ref{tab:top} that ChatGPT outputs similar sentences from different languages with similar structures.

In the Table~\ref{tab:top}, the left and right columns are the corresponding Chinese-English syntactic tree pairs with similar structure, among which, the DTED score of the first two pairs is 1, and the DTED score of the third pair is 0.8.
Obviously, although the sentences in the syntactic tree pairs are from different languages, the output of ChatGPT still has a high degree of similarity even though the wrong parsing outputsare given.

\section{Examples of English Linguistic Analysis}
\label{app:en:example}
\noindent\textbf{Predicative Verb-Centrism}: 
In the sentence "What if Google Morphed Into GoogleOS?", the correct root token should be "what". However, the ChatGPT result indicates that the subordinate clause is not recognized, and only the predicate verb is identified. Conversely, in cases where a clause contains a main clause with a predicate verb, such as "One of the pictures shows a flag that was found in Fallujah". ChatGPT does not incorrectly mark the root as unmarked.

\noindent\textbf{Noun Subject Preference}: 
For example, "One of the pictures shows a flag that was found in Fallujah.  In this sentence, the actual subject within the phrase "one of the pictures" is "one" and "pictures" functions as the noun being quantitatively modified. The words "one", "of", and "the" are dependent on "pictures" as well. However, in reality, the true subject should be "one", "of", and "the" collectively. The reason is that if the subject were solely "pictures", the verb "shows" would not be in the third person singular form. Consequently, the analysis of the sentence exhibits a flaw in terms of marking the subject accurately.

\noindent\textbf{Preposition-Case Ambiguity}: 
In the sentence "Compare the flags to the Fallujah one." the standard analysis designates "to" as a case-grammatical marker, which is dependent on "one." The word "one" serves as the object being compared, and the presence of "to" indicates that "one" is the direct object. However, ChatGPT only identifies "to" as a preposition, overlooking its role as a case-grammatical marker. Additionally, there is ambiguity in the phrase "the Fallujah one" that follows. As ChatGPT only labels "to" as a preposition, it may interpret it as "to the Fallujah," suggesting that "the" is the determiner for the compound noun "Fallujah one." However, "Fallujah" does not naturally form a compound noun with "one," and it is not possible to establish a dependency relationship among "the," "Fallujah," and "one."

\noindent\textbf{Auxiliary Verb Mislabeling}:
In the sentence " I'm staying away from the stock." the correct root should be "staying," while "am" functions as the auxiliary verb assisting in tense formation. However, there is a mislabeling where "am" is incorrectly marked as the root.
Similarly, in the sentence "He has denied this." the root should be "denied," but ChatGPT mistakenly identifies "has" as the root.
Moreover, in the sentence "It does seem that Iranians frequently make statements and then hide behind the lack of proof." the root should be "seem", but ChatGPT erroneously identifies "does" as the root.
These instances highlight inconsistencies in root identification by ChatGPT, where the actual root is mislabeled in favor of auxiliary verbs or other words in the sentence.

\noindent\textbf{Adjective Modifier Ambiguity}: 
In the sentence "The clerics demanded talks with local US commanders." ChatGPT tends to analyze the sentence in a way that suggests a dependency between the first modifier and the second modifier, and another dependency between the second modifier and the noun "commanders". Although this dependency may raise semantic concerns, it is syntactically acceptable. However, according to the gold standard annotations, it is "local" that is dependent on "commanders" and "US" has a separate dependency with "commanders". In other words, there is no direct relationship between "local" and "US" in the gold standard annotations.
\section{Statistics of Chinese Linguistic Analysis}
\label{app:statis.ctb5}
We count the number of sentences appearing in ChatGPT outputs for several types of linguistic analysis given in Section 6.2, as shown in the Table~\ref{tab:statis_ctb5}.
\section{Examples of Chinese Linguistic Analysis}
\label{app:zh:example}
\begin{table}[t]
    \centering
    \begin{tabular}{ccc}
    \toprule
         & snt\_number & snt\_percentage \\
    \midrule
        obj-dobj & $355$ & $35.86\%$ \\
        nmod-compound & $145$ & $14.65\%$\\
        punct root & $106$ & $35.86\%$\\
        nmod-dobj & $56$ & $10.71\%$\\
        nummod-dep & $26$ & $2.63\%$\\
    \bottomrule
    \end{tabular}
    \caption{The statistics of linguistic categories of CTB5.}
    \label{tab:statis_ctb5}
\end{table}

\begin{table}[t]
    \centering
    \scalebox{0.65}{
    \begin{tabular}{cccccc}
    \toprule
    ID & Word & Pred-Head & Pred-Rel & Gold-Head & Gold-Rel \\
    \midrule
    \multicolumn{2}{c}{\textbf{Obj-Dobj}}\\
    1 & \begin{CJK*}{UTF8}{gbsn}
            维阿里
    \end{CJK*} & 2 & nsubj & 2 & nsubj \\
    2 & \begin{CJK*}{UTF8}{gbsn}
            看好
    \end{CJK*}& 0 & root & 0 & root \\
    3 & \begin{CJK*}{UTF8}{gbsn}
            欧文
    \end{CJK*} & 2 & obj & 2 & dobj \\ 

    \midrule
    \multicolumn{2}{c}{\textbf{Quote Root}}\\
    1 & \begin{CJK*}{UTF8}{gbsn}
            “
    \end{CJK*} & 0 & root & 2 & punct \\
    2 & \begin{CJK*}{UTF8}{gbsn}
            光头
    \end{CJK*}&1&nsubj&4&appos \\
    3 & \begin{CJK*}{UTF8}{gbsn}
            ”
    \end{CJK*} & 1&punct&2&punct \\ 
    4 & \begin{CJK*}{UTF8}{gbsn}
            维阿里
    \end{CJK*} & 2&flat&6&nsubj \\ 
    5 & \begin{CJK*}{UTF8}{gbsn}
            现在
    \end{CJK*} & 6&advmod&6&nmod:tmod \\ 
    6 & \begin{CJK*}{UTF8}{gbsn}
            担任
    \end{CJK*} & 1&ccomp&0&root \\ 
    7 & \begin{CJK*}{UTF8}{gbsn}
            英格兰
    \end{CJK*} & 8&flat&10&nmod:assmod \\ 
    8 & \begin{CJK*}{UTF8}{gbsn}
            超级
    \end{CJK*} & 9&compound&9&amod \\ 
    9 & \begin{CJK*}{UTF8}{gbsn}
            联赛
    \end{CJK*} &10&nmod&10&compound:nn \\ 
    10 & \begin{CJK*}{UTF8}{gbsn}
            俱乐部
    \end{CJK*} & 6&obj&11&appos \\ 
    11 & \begin{CJK*}{UTF8}{gbsn}
            切尔西队
    \end{CJK*} & 10&nmod&13&nmod:assmod \\ 
    12 & \begin{CJK*}{UTF8}{gbsn}
            的
    \end{CJK*} & 11&case&11&case \\ 
    13 & \begin{CJK*}{UTF8}{gbsn}
            教练
    \end{CJK*} & 6&obj&6&dobj \\ 
    14 & \begin{CJK*}{UTF8}{gbsn}
            。
    \end{CJK*} & 1&punct&6&punct\\

    \midrule
    \multicolumn{2}{c}{\textbf{Compound-Nmod}}\\
    1 & \begin{CJK*}{UTF8}{gbsn}
            我
    \end{CJK*} & 4&nsubj&5&nsubj \\
    2 & \begin{CJK*}{UTF8}{gbsn}
            一点
    \end{CJK*}&3&det&5&advmod \\
    3 & \begin{CJK*}{UTF8}{gbsn}
            也
    \end{CJK*} & 4&advmod&5&advmod\\ 
    4 & \begin{CJK*}{UTF8}{gbsn}
            不
    \end{CJK*} & 5&advmod&5&neg \\ 
    5 & \begin{CJK*}{UTF8}{gbsn}
            怀疑
    \end{CJK*} & 0&root&19&dep \\ 
    6 & \begin{CJK*}{UTF8}{gbsn}
            欧文
    \end{CJK*} & 5&obj&15&nsubj \\ 
    7 & \begin{CJK*}{UTF8}{gbsn}
            将
    \end{CJK*} & 8&aux&15&advmod \\ 
    8 & \begin{CJK*}{UTF8}{gbsn}
            是
    \end{CJK*} & 5&ccomp&15&cop \\ 
    9 & \begin{CJK*}{UTF8}{gbsn}
            未来
    \end{CJK*} &10&compound:nn&10&dep\\ 
    10 & \begin{CJK*}{UTF8}{gbsn}
            几
    \end{CJK*} & 11&nummod&15&dep \\ 
    11 & \begin{CJK*}{UTF8}{gbsn}
            年
    \end{CJK*} & 8&obl&10&mark:clf \\ 
    12 & \begin{CJK*}{UTF8}{gbsn}
            内
    \end{CJK*} & 11&case&10&case \\ 
    13 & \begin{CJK*}{UTF8}{gbsn}
            真正
    \end{CJK*} & 14&amod&15&amod \\ 
    14 & \begin{CJK*}{UTF8}{gbsn}
            的
    \end{CJK*} & 11&nmod&13&mark\\ 
    15 & \begin{CJK*}{UTF8}{gbsn}
            巨星
    \end{CJK*} & 8&obj&5&ccomp\\ 
    16 & \begin{CJK*}{UTF8}{gbsn}
            ，
    \end{CJK*} & 5&punct&19&punct\\ 
    17 & \begin{CJK*}{UTF8}{gbsn}
            ”
    \end{CJK*} &5&punct&19&punct\\ 
    18 & \begin{CJK*}{UTF8}{gbsn}
            他
    \end{CJK*} & 20&nsubj&19&	nsubj\\ 
    19 & \begin{CJK*}{UTF8}{gbsn}
            说
    \end{CJK*} & 20	&ccomp	&0	&root\\ 
    20 & \begin{CJK*}{UTF8}{gbsn}
            。
    \end{CJK*} & 5	&punct	&19	&punct\\ 

    \midrule
    \multicolumn{2}{c}{\textbf{Error Root}}\\
    1 & \begin{CJK*}{UTF8}{gbsn}
            （
    \end{CJK*} & 0	&root&	5	&punct \\
    2 & \begin{CJK*}{UTF8}{gbsn}
            左
    \end{CJK*}& 1	&punct	&5	&dep \\
    3 & \begin{CJK*}{UTF8}{gbsn}
            一
    \end{CJK*} & 1	&punct&	5&	dep \\ 
    4 & \begin{CJK*}{UTF8}{gbsn}
            为
    \end{CJK*} & 1	&punct	&5	&dep \\
    5 & \begin{CJK*}{UTF8}{gbsn}
            作者
    \end{CJK*}& 4&	punct	&0	&root \\
    6 & \begin{CJK*}{UTF8}{gbsn}
            ）
    \end{CJK*} & 1	&punct	&5	&punct \\ 
    \midrule
    \multicolumn{2}{c}{\textbf{Dobj-Nmod}}\\
     1 & \begin{CJK*}{UTF8}{gbsn}
            发言人
    \end{CJK*} & 2&	nsubj&	2	&nsubj\\
    2 & \begin{CJK*}{UTF8}{gbsn}
            主张
    \end{CJK*}& 0	&root	&0	&root\\
    3 & \begin{CJK*}{UTF8}{gbsn}
            该国
    \end{CJK*} & 5&	nmod	&8	&nsubj \\ 
    4 & \begin{CJK*}{UTF8}{gbsn}
            就
    \end{CJK*} & 5&	advmod&	5	&case \\
    5 & \begin{CJK*}{UTF8}{gbsn}
            入侵
    \end{CJK*}& 2&	obj&	8	&nmod:prep\\
    6 & \begin{CJK*}{UTF8}{gbsn}
            邻国
    \end{CJK*} & 5&	nmod&	5	&dobj\\ 
    7 & \begin{CJK*}{UTF8}{gbsn}
            正式
    \end{CJK*}& 8&	advmod	&8&	advmod \\
    8 & \begin{CJK*}{UTF8}{gbsn}
            道歉
    \end{CJK*} & 2&	ccomp&	2	&ccomp\\ 
    \midrule
    \multicolumn{2}{c}{\textbf{Nummod-Dep}}\\
    1 & \begin{CJK*}{UTF8}{gbsn}
            十
    \end{CJK*} & 2	 &nummod &	3 &	dep\\
    2 & \begin{CJK*}{UTF8}{gbsn}
            面
    \end{CJK*}& 4 &	nsubj	 &1	 &mark:clf\\
    3 & \begin{CJK*}{UTF8}{gbsn}
            埋伏
    \end{CJK*} & 4 &	compound:nn	 &0	 &root \\ 
    4 & \begin{CJK*}{UTF8}{gbsn}
            ，
    \end{CJK*} & 2	 &punct &	3 &	punct \\
    5 & \begin{CJK*}{UTF8}{gbsn}
            创造
    \end{CJK*}& 2 &	conj	 &3	 &conj \\
    6 & \begin{CJK*}{UTF8}{gbsn}
            声势
    \end{CJK*} & 5 &	obj &	5 &	dobj\\
    \midrule
    \multicolumn{2}{c}{\textbf{Gold Errors}}\\
    1 & \begin{CJK*}{UTF8}{gbsn}
            内幕
    \end{CJK*} & 3	 &nsubj &	4 &	dep\\
    2 & \begin{CJK*}{UTF8}{gbsn}
            、
    \end{CJK*}& 1 &	punct	 &4 &punct\\
    3 & \begin{CJK*}{UTF8}{gbsn}
            或
    \end{CJK*} & 0 &root &4	 &cc \\ 
    4 & \begin{CJK*}{UTF8}{gbsn}
           丑闻
    \end{CJK*} & 3	 &conj &	0 &	root \\
    5 & \begin{CJK*}{UTF8}{gbsn}
            ？
    \end{CJK*}& 3 &	punct	 &4	 &punct \\ 
    \bottomrule
    
    \end{tabular}
    }
    \caption{The analysis examples in Chinese.}
    \label{tab:en.ex1}
\end{table}
\noindent\textbf{Obj-Dobj}: 
The most frequent label that differs between ChatGPT results and gold is "obj - dobj" (labeled obj in the outputs of ChatGPT and dobj in gold). In the CoNLL format data description provided, obj is the direct object relationship, while dobj (direct object) is also the direct object. And both obj and dobj appear in ChatGPT outputs, whereas there is no obj tag in gold. This suggests that the labels used in the two sets of annotation results are different, and that the labels in ChatGPT outputs are confusing. For example, in the following sentence, ChatGPT outputs and gold label the dependencies identically, but for the relationship between \begin{CJK*}{UTF8}{gbsn}
            "欧文"
\end{CJK*} and \begin{CJK*}{UTF8}{gbsn}
            "看好"
\end{CJK*} , ChatGPT outputs labels it as obj while gold labels it as dobj.

\noindent\textbf{Compound-Nmod}:
It is easier to judge the relationship of compound (compound noun) differently from gold, which is more often labelled as nmod, as in the example below, where \begin{CJK*}{UTF8}{gbsn}
            "联赛"
\end{CJK*} is dependent on \begin{CJK*}{UTF8}{gbsn}
            "俱乐部"
\end{CJK*} and ChatGPT labels the relationship as nmod, whereas gold is labelled as compound. The crucial aspect to consider is the presence of a conceptual overlap between "compound," which denotes a compound construction of a noun, adjective, or adverb, and "nmod," which signifies a nominal modifier, i.e., a noun, adjective, or adverb modifying another noun or pronoun. The distinction between these two categories is not always clearly defined, and there are instances where determining whether it should be labeled as "compound" or "nmod" can be subjective. Therefore, the labeling norms and conventions for these categories still remain a matter of debate and interpretation.

\noindent\textbf{Quote Root}:
When a sentence is spoken by someone and the speaker is positioned at the end of the sentence, resulting in a complete sentence with a predicate verb, the ChatGPT labeling scheme assigns the predicate verb as the root. On the other hand, the gold standard annotation assigns the speaker, represented by the Chinese character \begin{CJK*}{UTF8}{gbsn}
            "说"
\end{CJK*}, as the root in such cases. Similarly, in instances where ChatGPT outputs labels the verb\begin{CJK*}{UTF8}{gbsn}
            "怀疑"
\end{CJK*} as root, while gold labels \begin{CJK*}{UTF8}{gbsn}
            "他"
\end{CJK*} as root.
 These discrepancies in root labeling between ChatGPT outputs and gold exemplify the differing perspectives and criteria utilized in these annotation schemes.

\noindent\textbf{Error Root}:
The annotation of root in Chinese corpora can sometimes exhibit unexpected errors. One particular error involves the mislabeling of punctuation as the root by ChatGPT. Furthermore, ChatGPT does not consistently label predicate verbs as the root in all cases. These inconsistencies highlight the challenges and potential shortcomings in the annotation process for determining the root in Chinese sentences within the ChatGPT annotation scheme.

\noindent\textbf{Dobj-Nmod}: 
There are instances in which the ChatGPT labels the direct object (dobj) as a nominal modifier (nmod). One such case is exemplified by sentence \begin{CJK*}{UTF8}{gbsn}
            "发言人 主张 该国 就 入侵 邻国 正式 道歉"
\end{CJK*}, where ChatGPT considers \begin{CJK*}{UTF8}{gbsn}
            "邻国"
\end{CJK*} to be dependent on the noun \begin{CJK*}{UTF8}{gbsn}
            "入侵"
\end{CJK*} and assigns the relation as nmod. In contrast, the gold standard annotation marks the relation as dobj. This discrepancy in labeling suggests a disagreement in assigning the correct dependency relation between the two annotation schemes.

\noindent\textbf{Nummod-Dep}:
When it comes to number words in sentences, ChatGPT more frequently labels them as "nummod", whereas gold annotations often label them as "dep." This distinction arises primarily due to differing judgments regarding dependency relationships between the two annotation schemes. The variation in labeling can be attributed to differences in the interpretation of the role and dependency of number words within the sentence structure. It highlights the impact of subjective judgment in determining the appropriate dependency label for number words.

\noindent\textbf{Gold Errors}
Regarding the gold annotations, there is also a significant issue with errors in their labeling. In such cases, it becomes challenging to compare and analyze the preferences between gold and ChatGPT outputs. For example, in the given example, the words\begin{CJK*}{UTF8}{gbsn}"内幕"\end{CJK*}and \begin{CJK*}{UTF8}{gbsn}"丑闻"\end{CJK*}clearly have a coordinate relationship, and either of them can be considered as the root. The word \begin{CJK*}{UTF8}{gbsn}"或"\end{CJK*}should be identified as a coordinating conjunction, indicating the coordination between\begin{CJK*}{UTF8}{gbsn}"内幕"\end{CJK*}and \begin{CJK*}{UTF8}{gbsn}"丑闻"\end{CJK*}. If we assume \begin{CJK*}{UTF8}{gbsn}"或"\end{CJK*} as the root, and both \begin{CJK*}{UTF8}{gbsn}"内幕"\end{CJK*}and \begin{CJK*}{UTF8}{gbsn}"丑闻"\end{CJK*}as having an unknown relationship with \begin{CJK*}{UTF8}{gbsn}"或"\end{CJK*}, it would still make sense. However, in gold annotations, by labeling \begin{CJK*}{UTF8}{gbsn}"或"\end{CJK*} as a coordinating conjunction, and simultaneously marking the relationship between \begin{CJK*}{UTF8}{gbsn}"内幕"\end{CJK*} and \begin{CJK*}{UTF8}{gbsn}"丑闻"\end{CJK*} as unknown, it seems somewhat unreasonable.
In summary, the issue lies in the inconsistencies and potential errors within the gold annotations, making it challenging to establish a reliable basis for comparison and analysis against ChatGPT.


\end{document}